\documentclass[runningheads]{llncs}
\usepackage{amsmath,graphicx}
\usepackage{floatrow}
\usepackage{multirow}
\usepackage{hyperref}
\usepackage[misc]{ifsym} 

\usepackage{amsfonts}
\usepackage{comment}

\usepackage[T1]{fontenc}

\usepackage{graphicx}

\begin{document}
\title{Pose-based Tremor Classification for Parkinson’s Disease Diagnosis from Video}
\titlerunning{Pose-based Tremor Classification for PD Diagnosis from Video}

\author{Haozheng Zhang\inst{1}\orcidID{0000-0003-1312-4566} \and
Edmond S. L. Ho\inst{2}\orcidID{0000-0001-5862-106X} \and
Xiatian Zhang\inst{1}\orcidID{0000-0003-0228-6359} \and
Hubert P. H. Shum$^{(\textrm{\Letter})}$\inst{1}\orcidID{0000-0001-5651-6039}}

\authorrunning{H. Zheng et al.}
%
\institute{Durham University, UK \and Northumbria Univeristy, UK
\email{\{haozheng.zhang,xiatian.zhang,hubert.shum\}@durham.ac.uk\\e.ho@northumbria.ac.uk}}

\maketitle              

\begin{abstract}
Parkinson's disease (PD) is a progressive neurodegenerative disorder that results in a variety of motor dysfunction symptoms, including tremors, bradykinesia, rigidity and postural instability. The diagnosis of PD mainly relies on clinical experience rather than a definite medical test, and the diagnostic accuracy is only about 73-84\% since it is challenged by the subjective opinions or experiences of different medical experts. Therefore, an efficient and interpretable automatic PD diagnosis system is valuable for supporting clinicians with more robust diagnostic decision-making. To this end, we propose to classify Parkinson's tremor since it is one of the most predominant symptoms of PD with strong generalizability. Different from other computer-aided time and resource-consuming Parkinson's Tremor (PT) classification systems that rely on wearable sensors, we propose SPAPNet, which only requires consumer-grade non-intrusive video recording of camera-facing human movements as input to provide undiagnosed patients with low-cost PT classification results as a PD warning sign. For the first time, we propose to use a novel attention module with a lightweight pyramidal channel-squeezing-fusion architecture to extract relevant PT information and filter the noise efficiently. This design aids in improving both classification performance and system interpretability. Experimental results show that our system outperforms state-of-the-arts by achieving a balanced accuracy of 90.9\% and an F1-score of 90.6\% in classifying PT with the non-PT class.  

\end{abstract}
\section{Introduction}
Parkinson's disease (PD) is a progressive neurodegenerative disorder characterized by a variety of life-changing motor dysfunction symptoms, including tremor, bradykinesia (slow of movement), rigidity (limb stiffness), impaired balance and gait~\cite{patel2009}. According to pathological studies, the motor deficits of PD are mainly caused by the loss of dopamine due to the degeneration of dopamine neurons in patients~\cite{mhre2012}. As the second most common neurological disorder, the diagnosis of PD mainly relies on clinical criteria based on the parkinsonian symptoms (e.g., tremor, bradykinesia), medical history, and l-dopa or dopamine response~\cite{Gibb1988,Wirdefeldt2011,Mostafa2019}. However, the clinical diagnostic accuracy of PD is only about 73-84\%~\cite{Rizzo2016} since the diagnostic performance is challenged by the subjective opinions or experiences of different medical experts~\cite{Massano2012}. Therefore, an efficient and interpretable automatic PD diagnosis system is valuable for supporting clinicians with more robust diagnostic decision-making.

Recent machine learning and deep learning-based methods achieved impressive performance in PD diagnosis by analyzing the neuroimaging, cerebrospinal fluid, speech signals, 
gait pattern~\cite{Alle2021}, and hand tremors. Although neuroimagings~\cite{Zhang2020} or cerebrospinal fluid~\cite{Wangwu2020} based models perform well, they face a problem of high cost and intrusive. As for the non-intrusive methods, current speech-based models~\cite{Correa2019} are limited by their generalizability, as the language and pronunciation habits of people in different regions and countries vary significantly. Several studies~\cite{Hausdorff2009,Rizek2016} indicate that gait disturbance is less likely to be the main symptom in patients with early-onset PD, but more than 70\% of those patients present at least one type of tremors~\cite{Beitz2014,Pasquini2018,Rizek2016}. Hence we believe that detecting PD by diagnosing Parkinson's Tremor (PT) is a more generalizable approach compared with other methods. Conventional hand tremors-based studies~\cite{Hssayeni2019} achieve promising performance by using a deep learning network on wearable sensors data to detect PD. However, using wearable sensors is still time and resource-consuming~\cite{Hssayeni2019}, and requires careful synchronization of data captured from different sensors.

For the first time, we propose a graph neural network for diagnosing PD by PT classification as it effectively learns the spatial relationship between body joints from graph-structured data. Inspired by the information gain analysis~\cite{Li2020} and the clinician observation~\cite{Fahn2003} that PT usually occurs only on one side of the early stage PD patient’s upper body, we propose a novel attention module with a lightweight pyramidal channel-squeezing-fusion architecture to capture the self, short and long-range joint information specific to PT and filter noise. This design aids in improving both classification performance and system interpretability. Our system only requires consumer-grade non-intrusive video recordings and outperforms state-of-the-arts by achieving a balanced accuracy of 90.9\% and an F1-score of 90.6\% in classifying PT with non-PT class. Our work demonstrates the effectiveness and efficiency of computer-assisted technologies in supporting the diagnosis of PD non-intrusively, and provides a PT classification warning sign for supporting the diagnosis of PD in the resource-limited regions where the clinical resources are not abundant. Our source code is available at: \href{https://github.com/mattz10966/SPAPNet}{\color{blue} https://github.com/mattz10966/SPAPNet}.

\section{Method}

\begin{figure}
\includegraphics[width=0.98\textwidth]{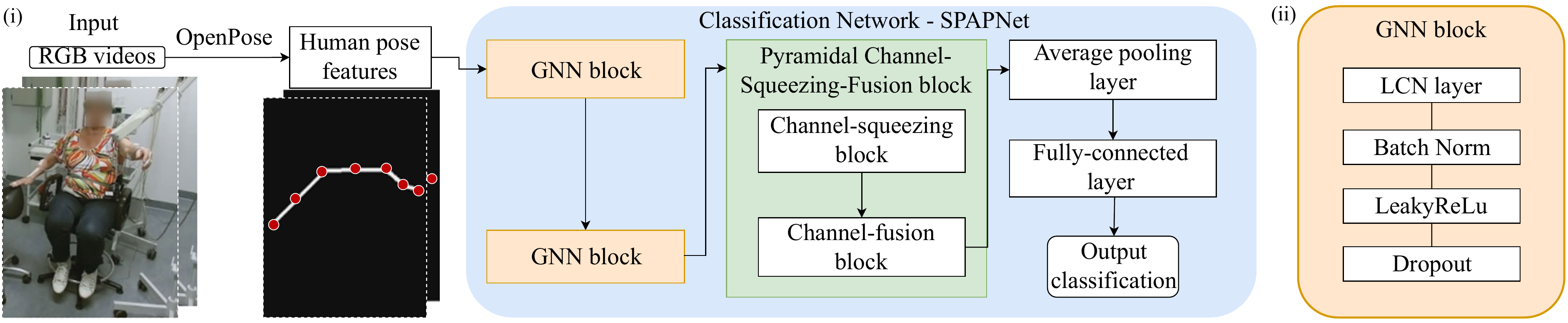}
\caption{(i) The overview of our proposed framework. (ii) The design of each GNN block.}
\label{net}
\end{figure}

As shown in Fig.~\ref{net}, the input consists of video recordings of each participant sitting in a chair in a normal upright position with various poses (e.g., tapping with the contralateral hand in the rhythm). We extract the human joint position features from the RGB video by OpenPose algorithm~\cite{cao2019OpenPose}. These human joint position features are passed to the Spatial Pyramidal Attention Parkinson's tremor classification Network (SPAPNet) for diagnosis. 

\subsection{Pose Extraction}
We first extract 2D skeleton features from the video sequences. Each frame is fed to OpenPose~\cite{cao2019OpenPose} due to its robust and efficient performance in detecting the 2D joint landmarks for people in normal upright positions. We do not estimate the 3D human pose as in \cite{Lu2020}, since the state-of-the-art 3D pose estimation methods still introduce noise while processing the 2D information to 3D~\cite{Chen2017,Luvizon2018,Wang2020}, which is not suitable for sensitive features like the tremor. We extract 18 OpenPose-skeleton format~\cite{cao2019OpenPose} landmarks with 2D coordinate $(x,y)$ and a confidence score $c$ indicating the estimation reliability by the OpenPose, but only use the seven upper body landmarks (seen in Fig.~\ref{fig3}) for PT classification, because PT usually tends to occur on the upper body, especially the hands and arms~\cite{Sveinbjornsdottir2016}. This approach eliminates less relevant features to help reduce model bias and improve efficiency. In addition, we do not include the head joint considering the participant's privacy, since the face is generally occluded in the medical video.
We implement normalization to reduce the bias from the underlying differences between the videos to tackle overfitting risk. To remove the participants' global translation, we center the participant's pose per frame by aligning the center of the triangle of the neck and two hip joints as the global origin. Then, we represent all joints as a relative value to the global origin.

\subsection{Classification Network}
We propose a \textit{Spatial Pyramidal Attention Parkinson's tremor classification Network (SPAPNet}) for PT diagnosis. The proposed SPAPNet consists of a graph neural network with the spatial attention mechanism and a novel pyramidal channel-squeezing-fusion block to enhance the attention mechanism.
\\
\\
\noindent \textbf{Graph Neural Network with Spatial Attention Mechanism:}
\\
\textit{Graph Neural Network (GNN):}
We propose to use the graph neural network to diagnose PD by classifying PT, since it effectively learns the spatial relationship between human joints from graph-structured data (e.g., human poses). To this end, we follow~\cite{Yan2018} to apply a pose graph \begin{math} G = (V,E)\end{math} aligned with the human skeletal graph to structure human pose data in the graph domain. In this graph, \begin{math} \{V = {v_{pq}}\}\end{math} denotes the joints positions, where \begin{math} v_{pq} \end{math} represents the \textit{p}-th 
joint at \begin{math} q\end{math}-th frame. The edge set \textit{E} includes: (1) the intra-skeleton connection each frame designed by the natural connections of human joints. (2) the inter-frame connections which connect the joints in consecutive frames.

\paragraph{Spatial Attention Mechanism: }
To improve the PT classification performance and interpret system the by human joints' importance, we propose using the spatial attention mechanism. Specifically, it interprets the important joints that the network considers in PT classification at each frame and video by attention weights and the temporal aggregation of the attention weights, respectively.

\begin{figure}
\includegraphics[width=\textwidth]{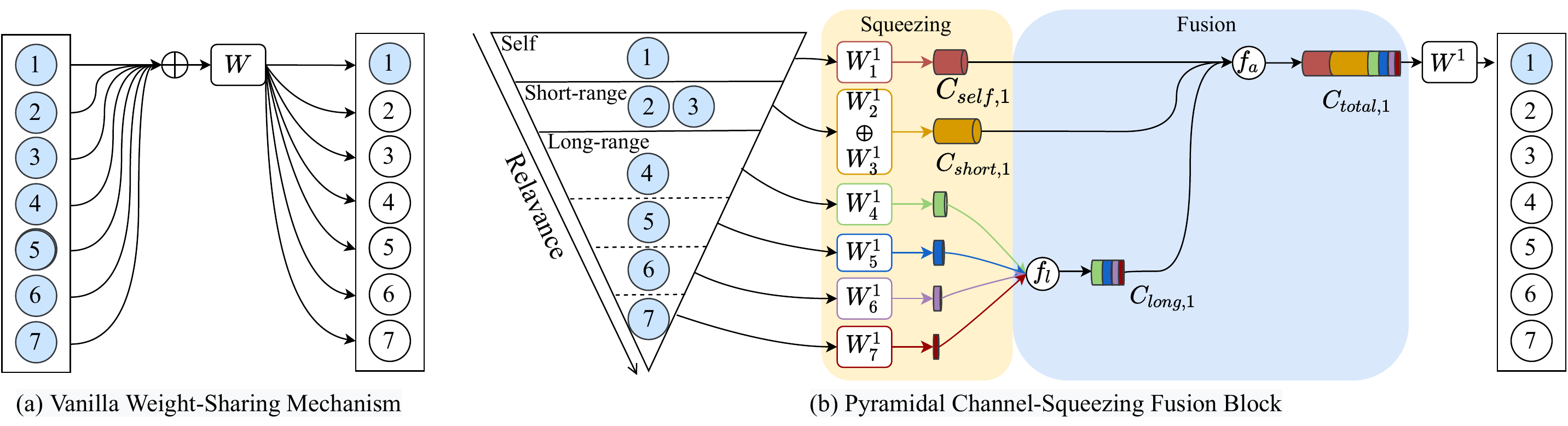}
\caption{The architectures of (a) Vanilla weight-sharing mechanism in GCN~\cite{Kipf2017}, (b) Proposed Pyramidal Channel-Squeezing-Fusion (PCSF) mechanism. Both architectures are taking the joint node 1, the right wrist as an example. Other nodes refer to Fig. \ref{fig3}.}
\label{fig1}
\end{figure}

We adopt the locally connected network (LCN)~\cite{Ci2022} to learn joint \textit{i}'s attention weight from its relationship between other joints. This method overcomes the representation power limitation that different joints share the same weight set in the vanilla graph convolutional network (GCN)~\cite{Kipf2017}. In addition, it enables the system to learn joint \textit{i}'s attention from its relationship between other joints. The basic formulation is as follows:

\begin{equation}
    \mathbf{h_{i}}= \sigma \left( \sum_{j\in \mathcal{N}^{i}} \mathbf{W}_{j}^{i} \mathbf{x}_j \hat{a}_{ij} \right)
\end{equation}
where $\mathbf{W}_{j}^{i}$ is the learnable attention weight between the target joint \textit{i} and the related joint \textit{j}, $\hat{a}_{ij}$ is the corresponding element in the adjacency matrix, $\mathbf{x}_j$ is the input features of node \textit{j}, $\mathcal{N}^{i}$ is the set of connected nodes for node \textit{i}, $\sigma$ is an activation function, and $\mathbf{h}_i$ is the updated features of node \textit{i}.
\\
\\
\noindent \textbf{Pyramidal Channel-Squeezing-Fusion Block (PCSF):}
As an extension of the spatial attention module, we propose a novel lightweight inverted pyramid architecture consisting of a \textit{channel-squeezing block} and a \textit{channel-fusion block} to 
extract relevant PT information and filter noise. This is motivated by two findings: (i) Information Gain analysis \cite{Li2020} shows that the information gain decreases exponentially with increasing distance between graph nodes; (ii) clinical observation \cite{Fahn2003} shows that PT usually occurs only on one side of the PD patient's upper body, such that the information relevancy between two arms should be reduced. Our proposed design does not require learnable parameters, such that it prevents overfitting problems. As illustrated in Fig. \ref{fig1}, we introduce the proposed PCSF by comparing it with the vanilla weight-sharing strategy in GCN~\cite{Kipf2017}. In PCSF, the final attention weight for joint-1 is learned from the information between the target joint 1 and the relevant joints 2,3,...,7 after a series of channel squeezing and fusion operations. Conversely, the vanilla weight-sharing mechanism can not learn from the joint-wise relevancy since all joints share the same set of weights.

\paragraph{The Channel-squeezing Block:}  To capture the relevant information specific to PT and filter noise, we hypothesize that (i) the short-range joints (i.e., on the same side of the body) contain slightly less relevant information compared with the target joint itself based on the information gain analysis; (ii) the long-range nodes (i.e., not on the same side of the body) contains much less information relevant to PT classification based on the clinician observation~\cite{Bhat2018,Fahn2003}. Hence, we propose the following channel-squeezing operation to reach the above hypothesis:

Suppose node \textit{m} to be the target node, node \textit{k} to be the relevant node of \textit{m}, such that the shortest path between two nodes in the graph domain is $k-a$. We propose Eq.\ref{sq} to determine the output channel size of the relevant node \textit{k}:
\begin{equation}
    C_{out,k} = b \times C_{in}, \quad \lvert k-m\rvert \leq 2 \quad and
    \quad
    C_{out,k} = d^{\lvert k-m\rvert}C_{in}, \quad \lvert k-m\rvert > 2
    \label{sq}
\end{equation}
where \textit{b}, \textit{d} are the channel-squeezing ratios for short-range and long-range node, respectively. $b,d\in[0,1]$ and $b\gg d$. $C_{out,k}$ is the output channel size of node \textit{k}. $\lvert \cdot \rvert$ is the distance between node \textit{m} and \textit{k} in the graph domain.

\paragraph{The Channel-fusion Block:} To fuse the relevancy information of the target joint \textit{m} from different ranges, we propose a two-stage fusion process to first fuse long-range features from less-related joints by $f_l$, then fuse all features by $f_a$:
\begin{equation}
    \mathbf{h_m}= {f_a} [\mathbf{h_{slef}},\mathbf{h_{short}},f_{l}(\mathbf{h_{long,p}}) \mathbf{W^m}
    \label{para}
\end{equation}
where $\mathbf{h_{long,p}}$ is features of long-range related node \textit{p}, $\mathbf{h_{short}}$ and $\mathbf{h_{slef}}$ are features of short-range related nodes and self-range node, respectively. $\mathbf{W^a}$ is the final weight of node \textit{m}.  

\subsubsection{Implementation Details:}
As shown in Fig. \ref{net}, we use two GNN blocks (64, 128 output channel size respectively) with each consisting of an LCN layer, a batch normalization layer, an LeakyReLU layer (0.2 alpha), and a dropout layer (0.2 rates). After two GNN blocks, we apply a PCSF block, a global average pooling layer and a fully connected layer. We use the focal-loss~\cite{Lin2017} as the loss function for overcoming class imbalance in multiclass classification task. The optimizer is chosen as Adam, and we train the model with a batch size of 16, a learning rate of 0.01 with 0.1 decay rate, and a maximum epoch of 500 for binary classification; For multiclass classification, the learning rate, weight decay, batch size, and epoch are 0.001, 0.1, 500, 8, and 500, respectively. Empirically, we set the short- and long-range channel-squeezing ratios \textit{b}, \textit{d} to 0.9 and 0.125, respectively, returns the most consistently good results.

\section{Experiments}
Our experiments were run on a PC with Ubuntu 18.04 and an NVIDIA GeForce RTX 3080. Our system is low-cost as it only requires an average GPU memory usage of 1.48 gigabytes for training. The total model training time on the TIM-TREMOR dataset is about ten hours, including human pose features extractions from RGB videos. It only takes about 48s for the PT classification of 1000 frames 30FPS video recording($\sim$33s), which can be employed in interactive-time diagnosis.

\subsubsection{The dataset:}
We verify our model on a publicly available TIM-TREMOR (Technology in Motion Tremor Dataset) dataset~\cite{Pintea2018}. The dataset consists of 917 video recordings from 55 participants sitting in a chair and performing a set of 21 tasks, and videos range from 18 seconds to 112 seconds. There are 579 videos that present different types of tremors, including 105 PT, 182 Essential Tremor (ET), 88 Functional Tremor (FT), and 204 Dystonic Tremor (DT) videos. Another 60 videos have no tremor during the assessment. The remaining 278 videos with ambiguous diagnosis results are labeled as ``Other''.

\subsubsection{Setup:}
We first eliminated inconsistent videos to avoid label noise, that is, (i) videos with motion tasks recorded only on a minor subset of participants; (ii) videos with ambiguous diagnosis label -``other''. Then, we clip each video into samples of 100 frames each, with the number of clips depending on the length of the successive video frames where the participant is not occluded by the interaction with the clinician. Each clip inherits the label of the source video and is considered an individual sample. A voting system~\cite{Lu2020,Lu2021} is employed to obtain the video-level classification results. This clipping-and-voting mechanism increases the robustness of the system and augments the sample size for training. We employ a 5-fold cross-validation to evaluate our proposed system.

To evaluate the generalizability of the proposed method, we validate our system not only on the binary classification (i.e., classify PT label with non-PT labels), but also on a more challenging multiclass classification task that classifies samples with five tremor labels (PT, ET, FT, DT, and No tremor). We report the mean and standard deviation among all cross-validation for the following metrics: the metrics for the binary classification includes the accuracy (AC), sensitivity (SE), specificity (SP), and F1-Score; the metrics for the multiclass classification are AC and per-class and macro average F1-score, SE and SP.

\begin{table}
\label{tab1}

\scriptsize
\killfloatstyle
\ttabbox
{
\begin{tabular}{|c|c|c|c|c|c|}
\hline
\multicolumn{2}{|c}{ } &
\multicolumn{4}{|c|}{Binary Classification} 
\\
\hline
\multicolumn{2}{|c|}{Method } &
AC & 
SE &
SP &
F1 
\\
\hline
\multicolumn{2}{|c|}{CNN-LSTM~\cite{Wang2021} } &
81.0&
n/a &
79.0&
80.0 
\\
\multicolumn{2}{|c|}{LSTM~\cite{Wang2021} } &
80.0 &
n/a &
79.0 &
79.0 
\\
\multicolumn{2}{|c|}{SVM-1~\cite{Wang2021} } &
53.0 &
n/a &
63.0 &
55.0 
\\
\multicolumn{2}{|c|}{ST-GCN\cite{Yan2018} } &
87.7 $\pm$ 3.8  &
 88.3 $\pm$ 5.3  &
 87.4 $\pm$ 3.1  &
 87.0 $\pm$ 4.4   
\\
\hline
\multicolumn{2}{|c|}{CNN-Conv1D } &
 81.6 $\pm$ 5.7  &
 83.4 $\pm$ 9.1  &
 80.7 $\pm$ 4.4  &
 80.3 $\pm$ 6.0  
\\
\multicolumn{2}{|c|}{Decision Tree } &
 74.5 $\pm$ 4.7  &
 73.4 $\pm$ 5.7  &
 75.8 $\pm$ 4.0  &
 73.6 $\pm$ 4.6    
\\
\multicolumn{2}{|c|}{SVM  } &
 64.3 $\pm$ 5.4  &
 62.2 $\pm$ 7.5  &
 66.7 $\pm$ 4.6  &
 63.1 $\pm$ 7.1  
\\
\hline
\parbox[t]{2mm}{\multirow{3}{*}{\rotatebox[origin=c]{90}{Ours}}}&
SPAPNet - full &
 \textbf{90.9 $\pm$ 3.4}  &
 \textbf{90.7 $\pm$ 5.0 }  &
 \textbf{91.3 $\pm$ 2.3}  &
 \textbf{90.6 $\pm$ 3.7}   
\\
&
w/o PCSF &
 88.4 $\pm$ 4.5  &
 90.4 $\pm$ 6.9  &
 87.0 $\pm$ 3.7  &
 87.5 $\pm$ 5.2  
\\
&
w/o Attention &
 82.6 $\pm$ 5.3  &
 82.7 $\pm$ 6.0  &
 82.8 $\pm$ 5.1  &
 81.3 $\pm$ 6.8   
\\

\hline
\multicolumn{2}{|c}{ } &
\multicolumn{4}{|c|}{Multiclass Classification} 

\\
\hline
\multicolumn{2}{|c|}{ST-GCN [31] } &
 70.3 $\pm$ 6.9 &
 69.5 $\pm$ 6.4 &
 90.7 $\pm$ 5.4 &
 67.9 $\pm$ 6.7
\\
\hline
\multicolumn{2}{|c|}{CNN-Conv1D } &
 63.1 $\pm$ 6.5 &
 59.5 $\pm$ 5.6 &
 90.8 $\pm$ 7.4  &  
 61.9 $\pm$ 8.3  
\\
\multicolumn{2}{|c|}{Decision Tree } &
 54.3 $\pm$ 5.7 &
 49.0 $\pm$ 7.3 &
 92.3 $\pm$ 5.4  &
 55.5 $\pm$ 6.5 
\\
\multicolumn{2}{|c|}{SVM  } &
 47.6 $\pm$ 6.4 &
 45.7 $\pm$ 6.9 &
 91.6 $\pm$ 6.1 &
 52.1 $\pm$ 7.2  
\\
\hline
\parbox[t]{2mm}{\multirow{3}{*}{\rotatebox[origin=c]{90}{Ours}}}&
SPAPNet - full &
 \textbf{ 73.3 $\pm$ 6.8} &
 \textbf{ 72.8 $\pm$ 5.1}  &
 \textbf{ 92.3 $\pm$ 4.1    }  &
\textbf{ 70.7 $\pm$ 6.5} 
\\
&
w/o PCSF &
 69.1 $\pm$ 6.9 &
 69.9 $\pm$ 4.0  &
 88.2 $\pm$ 4.6  &
 65.7 $\pm$ 7.1
\\
&
w/o Attention &
  65.9 $\pm$ 6.8 &
 64.2 $\pm$ 5.5   &
 90.4 $\pm$ 7.9  &
 65.0 $\pm$ 7.9 
\\
\hline
\end{tabular}}
{\caption{The comparisons on the binary classification (PT v.s. non-PT) task and the summarized multiclass classification (PT v.s. ET v.s DT v.s FT v.s non-tremor) results.}\label{tab1}
}
\end{table}
\subsubsection{Comparison with Other Methods:}
To evaluate the effectiveness of our system,  we compare our results with the following state-of-the-art video-based PT classification methods:  (i) CNN-LSTM~\cite{Wang2021}: This method uses a CNN-LSTM model to classify the PT and non-PT classes from hand landmarks extracted by MediaPipe~\cite{Zhang2020a}, their data is videos from the TIM-TREMOR dataset; (ii) SVM-1~\cite{Wang2021}: This is a support vector machine model proposed to classify the PT and non-PT classes by the same features in \cite{Wang2021};  (iii) LSTM~\cite{Wang2021}: This is an LSTM deep neural network proposed to classify the PT and non-PT classes by the same features in \cite{Wang2021}; (iv) ST-GCN~\cite{Yan2018}: This is a spatial and temporal graph convolutional neural network for classification tasks on human pose data. For works in~\cite{Wang2021}, we only report the performance in their work since the source code is not publicly available. To compare the effectiveness of our system with conventional methods, we implement a CNN with 1D convolutional layers (CNN-Conv1D)~\cite{Wang2021} and two machine learning-based methods, namely Decision Tree (DT) and SVM. 

From the binary classification result in Table \ref{tab1}, our full system outperforms state-out-of-the-arts~\cite{Wang2021,Yan2018} and other implemented methods. Our AC, SE, SP, and F1 achieves over 90\% with standard deviations less than 5\%, which indicates the effectiveness and robustness in classifying PT class with non-PT class. Our system achieves better performance by only applying spatial convolution instead of a more deep architecture like spatial-temporal convolution modeling method, ST-GCN~\cite{Yan2018}. The result validates that our proposed PCSF block effectively improves classification performance and mitigates the overfitting risk in small datasets. Moreover, although our system is designed for binary classification purposes, the full system also shows effectiveness and generalizability by outperforming others in the multiclass classification task. The high macro-average SP showed relatively reliable performance in identifying people without corresponding tremor labels. Improving the multiclass classification AC and SE is scheduled in our future work.

\subsubsection{Ablation Studies:}  We perform an ablation to evaluate whether there is any adverse effect caused by the proposed PCSF block or the whole attention module. From the rows of "Ours" in Table \ref{tab1}, we observe the effectiveness of the PCSF block and attention module from the performance reduction across all metrics when eliminating the PCSF  or the whole attention module for both classification tasks. In addition, we observe the stability of using the full system as it has smaller standard deviations than its variants. Besides, we can observe that the vanilla GNN (i.e., SPAPNet w/o Attention) presents better performance than CNN-Con2D in both classification tasks. It demonstrates the effectiveness of learning human pose features in the graph domain. Moreover, the results show the advantage of deep learning networks by comparing them with two machine learning-based methods, which are decision tree and SVM.

\subsubsection{Qualitative Analysis:}  
\begin{figure}
\includegraphics[width=0.9\textwidth]{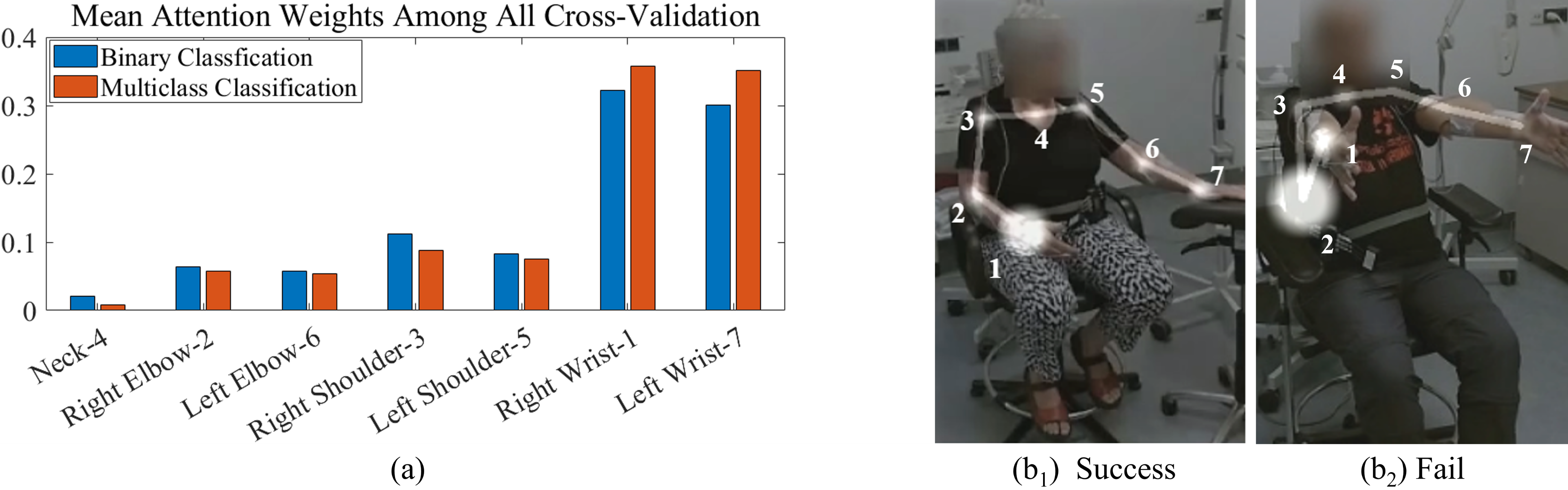}

\caption{(a) The mean attention weights of different joints among all cross-validation for both classification tasks; (b) The visualization of the attention weights at a single example frame. The joint index numbers in (b) corresponds to (a); (b$_1$) One frame in a successful diagnosis; (b$_2$) One frame in a false diagnosis. } \label{fig3}
\end{figure}

Fig.~\ref{fig3}a. visualizes the interpretability of our system by presenting the mean attention weights of each skeleton joint among all cross-validation. We notice that the mean attention weights of `Right Wrist' and `Left Wrist' are significantly higher than others on both classification tasks. It indicates our system pays more attention to the movements of participants' wrists. In addition, the attention weight of `Neck' is lower than others significantly. One possible reason is that the participants are sitting on the chair, and their neck joint has the smallest global variance during the whole video.

We also analyze the situation in which our method fails or succeeds. Fig.~\ref{fig3} b$_1$. is a frame in a successful diagnosed example of a PT patient. Consistent with the clinician PT diagnosis based on right hand resting tremor, the right wrist node contributes the most attention. Fig.~\ref{fig3} b$_2$. is a frame in misdiagnosis, and the attention is incorrectly dominated by the mis-detected joint position of the right elbow from the pose extraction algorithm. Therefore, it highlights the importance of improving pose extraction performance for future work. 

\section{Conclusion}
In this work, we propose a novel interpretable method SPAPNet to diagnose Parkinson's from the consumer-grade RGB video recordings. Our system outperforms state-of-the-arts by achieving an accuracy of 90.9\% and an F1-score of 90.6\%. The proposed attention module aids in improving both classification performance and system interpretability. Our proposed novel lightweight pyramidal channel-squeezing-fusion block effectively learns the self, short and long-range relevant information specific to Parkinson's tremor and filters irrelevant noise. Our system shows the potential to support non-intrusive PD diagnosis from human pose videos. Since our system only requires the consumer-grade human pose videos as input, it provides a way for diagnosis of PD in the resource-limited regions where the clinical experts are not abundant. In addition, our system shows potential for remote diagnosis of PD in special situations (e.g., COVID-19 epidemic) and automatic monitoring of PT symptoms during daily life for PD diagnosis.

%
%
%
%

\end{document}